\documentclass[conference]{IEEEtran}
\IEEEoverridecommandlockouts
\usepackage{cite}
\usepackage{amsmath,amssymb,amsfonts}
\usepackage{algorithmic}
\usepackage{graphicx}
\usepackage{textcomp}
\usepackage{xcolor}
\usepackage{hyperref}
\def\BibTeX{{\rm B\kern-.05em{\sc i\kern-.025em b}\kern-.08em
    T\kern-.1667em\lower.7ex\hbox{E}\kern-.125emX}}

\begin{document}

\title{Pose Estimation of Buried Deep-Sea Objects using 3D Vision Deep Learning Models \\
\thanks{$^{\ast}$Denotes equal contribution. \\
NOAA Award: NA20OAR4320278 Southern California DDT ocean dumpsite characterization, monitoring, and research pilot project and Office of Naval Research award N00014-22-C-2006 Marine Robotics Testbed}
}

\author{\IEEEauthorblockN{Jerry Yan$^{\ast 1,2}$, Chinmay Talegaonkar$^{\ast 2}$,  Nicholas Antipa$^2$, Eric Terrill$^1$, Sophia Merrifield$^1$ }
\IEEEauthorblockA{$^1$\textit{Marine Physical Laboratory},
\textit{Scripps Institution of Oceanography, UCSD}, La Jolla, CA USA \\
$^2$\textit{Department of Electrical and Computer Engineering},
\textit{UCSD}, La Jolla, CA USA \\
}}

\maketitle

\begin{abstract}
We present an approach for pose and burial fraction estimation of debris field barrels found on the seabed in the Southern California San Pedro Basin. Our computational workflow leverages recent advances in foundation models for segmentation and a vision transformer-based approach to estimate the point cloud which defines the geometry of the barrel. We propose BarrelNet for estimating the 6-DOF pose and radius of buried barrels from the barrel point clouds as input. We train BarrelNet using synthetically generated barrel point clouds, and qualitatively demonstrate the potential of our approach using remotely operated vehicle (ROV) video footage of barrels found at a historic dump site. We compare our method to a traditional least squares fitting approach and show significant improvement according to our defined benchmarks.
\end{abstract}


\section{Introduction}
Image-based 3D modeling and analysis of submerged man-made objects is an important tool for the inspection of seabed infrastructure and cultural preservation. Our study focuses on man-made objects imaged using a remotely operated vehicle (ROV) operated in the Southern California San Pedro Basin during a 3 week cruise to conduct a wide area survey of a historic dump site \cite{merrifield_wide-area_2023}. Historical dumpsites, oceanic locations where hazardous chemicals were discarded via bulk and containerized methods, pose risks to the marine food web and human health \cite{mackintosh_newly_2016}. The region of our survey is a known dumpsite for the chemical dichlorodiphenyltrichloroethane (DDT) which has been detected extensively in the basin's sediment \cite{schmidt_disentangling_2024}, and has a large number of distributed, barrel-sized ($\sim$1m) objects \cite{merrifield_wide-area_2023}.
The majority of existing research on seabed debris analysis focuses on the detection and classification of objects from images, but the development of automated techniques to estimate pose and burial fraction has been limited. Underwater object recognition and pose estimation from 3D point clouds typically requires accurate scans (usually laser) and an unobstructed view of the object to accurately reconstruct the 3D point cloud \cite{himri_3d_2019}. However, objects buried in the seabed are only partially visible, complicating accurate pose estimation from conventional methods. The collected footage is affected by unstable ROV movements and challenging deep-sea conditions, such as haze, marine snow, low contrast, and loss of detail \cite{li_underwater_2019}. These factors make conventional photogrammetric reconstruction methods difficult, as they typically require stable imaging conditions similar to those found in land-based datasets. This serves as a motivation for creating a new robust framework for pose and burial fraction estimation of objects buried in the sea-bed.

We propose a learning-based approach that overcomes the data quality limitations of the ROV footage to predict the pose and burial fraction of barrels from images captured at different points in the ROV flight trajectory. Our method leverages a combination of learning-based 3D reconstruction and foundation segmentation models. For pose and burial fraction estimation, we introduce \textit{BarrelNet}, a modification of the PointNet architecture \cite{qi_pointnet_2017}, which we train on synthetically generated cylinder point clouds. These generated barrel point clouds account for occlusion from the camera's viewpoint and the burial of barrels in the seafloor. We evaluate our workflow on synthetic test data against a classical cylinder fitting method and show significant improvement. We also demonstrate qualitative generalization of the barrel poses and burial fraction on real imagery collected by the ROV.

\section{Related Work}
Our algorithm is inspired by modern learning-based approaches for multi-view 3D reconstruction and 6-DOF pose estimation. Below we briefly outline the relevant literature from these fields: 

\subsection{Underwater 3D Reconstruction}
The industry-standard method for multi-view 3D reconstruction is photogrammetry, which jointly estimates a 3D point cloud of the scene and the camera extrinsics and intrinsics using structure from motion (SFM). Photogrammetry has been extensively used for terrestrial and aerial surveys \cite{agarwal_building_2011, svennevig_oblique_2015,schonberger_structure--motion_2016}, and methods have also been developed for archaeological and ecological underwater surveys \cite{arnaubec_underwater_2023, sedlazeck_3d_2009, pizarro_large_2004, drap_underwater_2013}.

We leverage the recent advances in learning-based approaches for 3D reconstruction and geometry estimation \cite{smith_flowmap_2024, wang_dust3r_2024, wang_visual_2023}.
Unlike conventional SFM methods, these algorithms do not rely on heuristic-based steps like feature detection and matching. They are more robust to erratic or poor camera movement. The resulting point clouds are less noisy and have fewer gaps compared to conventional photogrammetry tools like COLMAP (\autoref{fig:colmap-vs-dust3r}).

\subsection{6-DOF Pose Estimation Methods}
6-DOF pose estimation can predominantly be divided into image and point cloud based approaches. Image based methods \cite{wen_foundationpose_2024, li_nerf-pose_2023, peng_pvnet_2018, park_pix2pose_2019, rekavandi_b-pose_2023} have been adopted for underwater 6-DOF pose estimation of cameras on AUVs \cite{nielsen_evaluation_2019}, but not for objects in the ocean. Most methods require keypoint annotations on CAD models, which will not be reliable for barrels with different textures resulting from biological growth and/or corrosion as observed in real data. We therefore focus on point cloud-based methods.

Classical methods \cite{schnabel_efficient_2007, zhang_fast_2021} rely on point cloud registration or iterative refinement, which are sensitive to initialization, occlusion, and noise in the data. In the case of fitting cylinders to point clouds, classical methods generally only consider points on the side of a cylinder and exclude the cap points, meaning the inclusion of caps in the point cloud will impact the fitting accuracy. The simplest way to estimate a cylinder axis is to estimate the point cloud surface normals and estimate least squares on the normals since the axis will always be orthogonal to them \cite{vosselman_recognising_2003, schnabel_efficient_2007}. Newer methods are more robust \cite{nurunnabi_robust_2017}, but all rely on outlier detection like RANSAC \cite{fischler_random_1981} to produce good results. This means if cap points are included, RANSAC must ignore them for a good fit, but since the caps on barrels are almost always visible and make up a non-negligible fraction of points, it leads to inconsistent results.

Recent learning-based models have shown improved performance in pose accuracy directly from point clouds \cite{li_supervised_2019, ge_hand_2018} using the PointNet \cite{qi_pointnet_2017} architecture. These approaches are tested mostly on synthetic datasets, and suffer in performance in case of occlusions and partially visible point clouds \cite{wang_unsupervised_2021}. Our proposed approach handles partial visibility of point clouds. This is enabled by generating synthetic data to correctly model occlusions resulting from burial in ocean floor as well as self-occlusion from cameras. We qualitatively show that our approach generalizes to real data. Our work extends a PointNet based approach for this task, to estimate both burial fraction, and axis of the barrel.

\section{Problem Formulation}
Given seabed imagery from a moving ROV, we wish to estimate the pose and burial fraction ($b_f$) of partially buried barrels visible in the footage. 
We assume that each video footage contains only a single visible barrel. The pose of a cylindrical barrel can be parameterized by a unit vector $\vec{\mathbf{n}}$ along the barrel axis, and $\vec{\mathbf{c}}$, which denotes the centroid of the barrel. Since we assume the barrel is symmetric along its axis, the barrel pose has five degrees of freedom. To avoid ambiguity, we always assume barrel axis pointing outward from the ocean floor. We assume a right handed $z$-up coordinate system, with $xy$ plane being the ocean floor. Let $r$ denote the radius of the barrel. Using prior knowledge for the collected dataset, we infer the height of the barrel $h$ using the known ratio of $\frac{r}{h}$ as barrel dimensions are standardized, even when buried.

Our method predicts $\vec{\mathbf{n}}, r, \vec{\mathbf{c}}$ for the barrel observed in the input video footage. The burial fraction is independent of $h$, so we assume $h=1$, and predict the radius as a fraction of $h$. This is consistent with real data, that $r<h$ for the observed barrels. The predictions from our method are used to instantiate the barrel in the estimated pose.  We then calculate the burial fraction via Monte Carlo sampling of a sufficiently large number of points $N$ points inside this cylinder's volume.
\begin{equation}
    b_f = \frac{N_{z\leq 0}}{N},
    \label{eqn:1}
\end{equation}
where $N_{z\leq0}$ is the number of sampled points with $z$ coordinate less than 0.

\begin{figure*}[!ht]
    \centering
    \includegraphics[width=\linewidth]{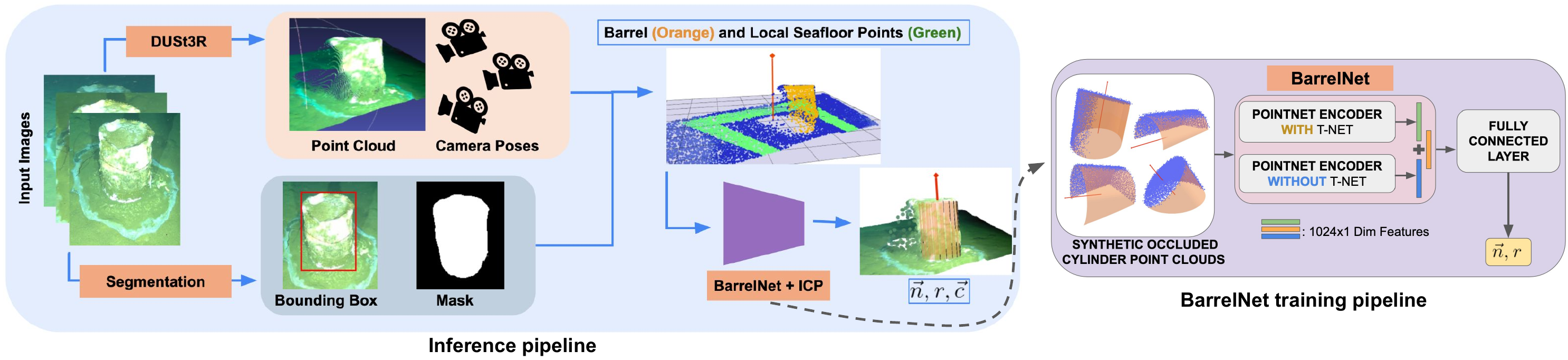}
    \caption{The overall processing workflow for barrel pose estimation. Given a series of underwater images, we reconstruct the scene as a 3D point cloud using DUSt3R \cite{wang_dust3r_2024} and mask the barrel using Grounding DINO + Segment Anything \cite{liu_grounding_2023, kirillov_segment_2023}. We then use this information to isolate the barrel in the point cloud, which we then feed to BarrelNet (a modification to the PointNet \cite{qi_pointnet_2017} architecture) to find the cylinder axis and its dimensions.}
    \label{fig:Pipeline_Figure}
\end{figure*}

\section{Method}
Our method broadly consists of the following steps: From the set of $M$ images $(\mathbb{R}^{W\times H})$ in the video, we first estimate camera orientation and generate a 3D point cloud of the scene $P_{full}$ using DUSt3R. We then apply Grounding DINO \cite{liu_grounding_2023, zhang_dino_2022} for identifying bounding boxes around the targets of interest and Segment Anything Model (SAM) \cite{kirillov_segment_2023} to get masks of the barrels in the images. Using the masks and bounding boxes, the point cloud of the barrel $P_{barrel}$ and the ocean floor $P_{floor}$ around it is extracted from the entire scene $P_{full}$. 
We feed $P_{barrel}$ as input to the proposed BarrelNet to estimate $\vec{\mathbf{n}}, r$, which we can use to generate a predicted barrel point cloud $P'_{barrel}$. To predict $\vec{\mathbf{c}}$, we use simplified iterative closest point (ICP) \cite{besl_method_1992} to align $P'_{barrel}$ to $P_{barrel}$. From the final outputs, the burial fraction is computed using \autoref{eqn:1}. The workflow components are explained in the subsections below.

\subsection{DUSt3R for Barrel Site 3D Reconstruction}
\begin{figure}[h!]
    \centering
    \includegraphics[width=0.7\linewidth]{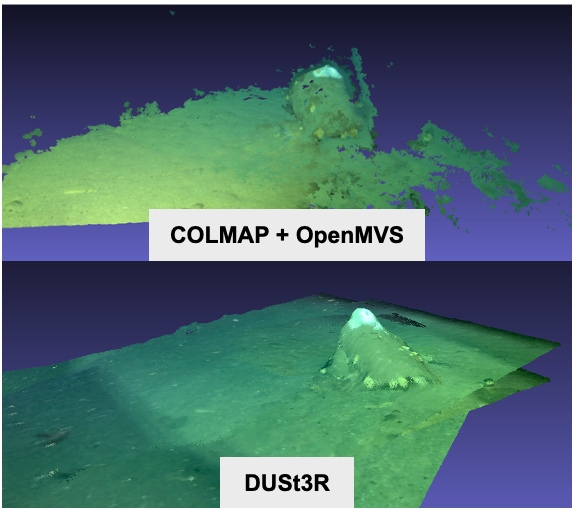}
    \caption{A qualitative comparison from roughly the same viewpoint between 3D reconstructions from COLMAP + OpenMVS (top) and DUSt3R (bottom). COLMAP + OpenMVS is highly susceptible to imaging conditions, leading to noisy and incomplete point clouds, while DUSt3R creates smooth (sometimes too smooth) and complete point clouds.}
    \label{fig:colmap-vs-dust3r}
\end{figure}

DUSt3R \cite{wang_dust3r_2024} leverages a pre-trained vision transformer \cite{dosovitskiy_image_2021} architecture to predict pointmaps $\{p_{i,j}\in\mathbb{R}^3:i=1,\dots,W,j=1,\dots,H\}$ and camera poses for each image/camera viewpoint. It performs global alignment between the predicted pointmaps $p_{i,j}$, to produce $P_{full}$ and in the process also estimates camera orientations. DUSt3R is robust to small dynamic motion noise in the water column like marine snow and jellyfish. The resulting point cloud is smoother and has lesser holes compared to COLMAP (\autoref{fig:colmap-vs-dust3r}). While DUSt3R is a very promising approach, it is not designed for the shortcomings of underwater imagery. Contrast enhancement of images using CLAHE \cite{pizer_adaptive_1987} improves performance in most cases. Additionally, instead of evenly sub-sampling frames, we use a contiguous subset of frames as input, as we observed that DUSt3R camera pose estimation suffers if the viewpoints differ too greatly from each other.

\subsection{Barrel Segmentation and Point Cloud Extraction}
Images are provided to transformer-based object detector Grounding DINO \cite{liu_grounding_2023, zhang_dino_2022} with the label ``underwater barrel" to extract the bounding box around the barrel. This bounding box is fed to SAM to reduce the image search area and extract a mask of the barrel (\autoref{fig:Pipeline_Figure}), which we will denote as $M_{barrel}^{\circ}$. We extract the barrel region bounding box and the local seafloor around it as follows: Let $\alpha$ be a scaling factor multiplied to the original bounding box. $B_{\alpha}$ denotes the region encompassed by the scaled bounding box. Ocean floor around the barrel lies in the region $M_{floor}=B_{\alpha_1} \setminus B_{\alpha_2}, \alpha_1 > \alpha_2 > 1$. To ensure no stray points from the ocean floor are extracted as part of the barrel point cloud $P_{barrel}$, we erode \cite{vizilter2015morphological} the original barrel mask so that $M_{barrel}=M_{barrel}^{\circ}\ominus 1_{5\times 5}$. Our barrel and floor masks are used to extract $P_{barrel}$ and $P_{floor}$ respectively from each view by projecting $P_{full}$ into the full image field of view and masking the points accordingly. Since DUSt3R generates the point cloud in an arbitrary coordinate frame, we estimate the normal vector $\vec{\mathbf{n}}_{floor}$ to the ocean floor around the barrel by fitting a plane to the points in $P_{floor}$. $P_{full}$ is then rotated and translated such that $\vec{\mathbf{n}}_{floor}$ aligns with the positive $z$-axis, and the centroid of $P_{floor}$ is on the $z=0$ plane (\autoref{fig:Pipeline_Figure}).

\subsection{BarrelNet and ICP: Estimating Barrel Parameters}
BarrelNet modifies the PointNet \cite{qi_pointnet_2017} architecture used for $k$-class classification. We set $k=4$ to predict $v\in\mathbb{R}^4$. PointNet architecture consists of a T-Net module which learns how to rotate input point clouds to make the final PointNet features rotation invariant. This provides rotational invariance to PointNet for classification. But in our case $r$ is independent of rotation about $z$-axis, and $\vec{\mathbf{n}}$ is not. Hence, we use a combination of rotation equivariant and invariant features which allows our model to predict both rotation invariant and equivariant parameters. By disabling T-Net, the resulting feature vector is rotation equivariant. We use a sigmoid activation function ($\sigma(x)=\frac{1}{1+e^{-x}}$) on the first element of $v$ to predict $r$. The last 3 elements of $v$ are normalized to a unit vector with positive $z$ coordinate to predict $\vec{\mathbf{n}}$. \autoref{fig:Pipeline_Figure} illustrates the BarrelNet training pipeline in more detail. 

To estimate the burial fraction ($b_f$) of the Barrel, we need to estimate the barrel centroid location $\vec{\mathbf{c}}$. We use a simplified version of ICP to align $P_{barrel}$ and $P'_{barrel}$. ICP aligns two point clouds by iteratively finding the closest points and estimating the transformation (rotation and translation) between them. It then applies this transformation and repeats this process until convergence. In our case, BarrelNet predicts the axis of the barrel $\vec{\mathbf{n}}$. This allows us to keep the rotation fixed during ICP and optimize only for the translation of the centroid location $\vec{\mathbf{c}}$. Fixing the rotation of the barrel greatly reduces the search space, and allows consistently correct convergence with randomly sampled initializations of $\vec{\mathbf{c}}$. We run ICP 9 times, with each $\vec{\mathbf{c}}$ initialized in a sphere around the mean of the observed barrel point cloud.

The synthetic data to train this model is generated by sampling cylinders with different poses and burial percentages. Depth for each instance is rendered from 20 different camera views and is projected into 3D space to create synthetic point clouds, which are used as training data (\autoref{fig:Pipeline_Figure}).

\section{Experiments}

\begin{figure*}[ht!]
    \centering
    \includegraphics[width=\linewidth]{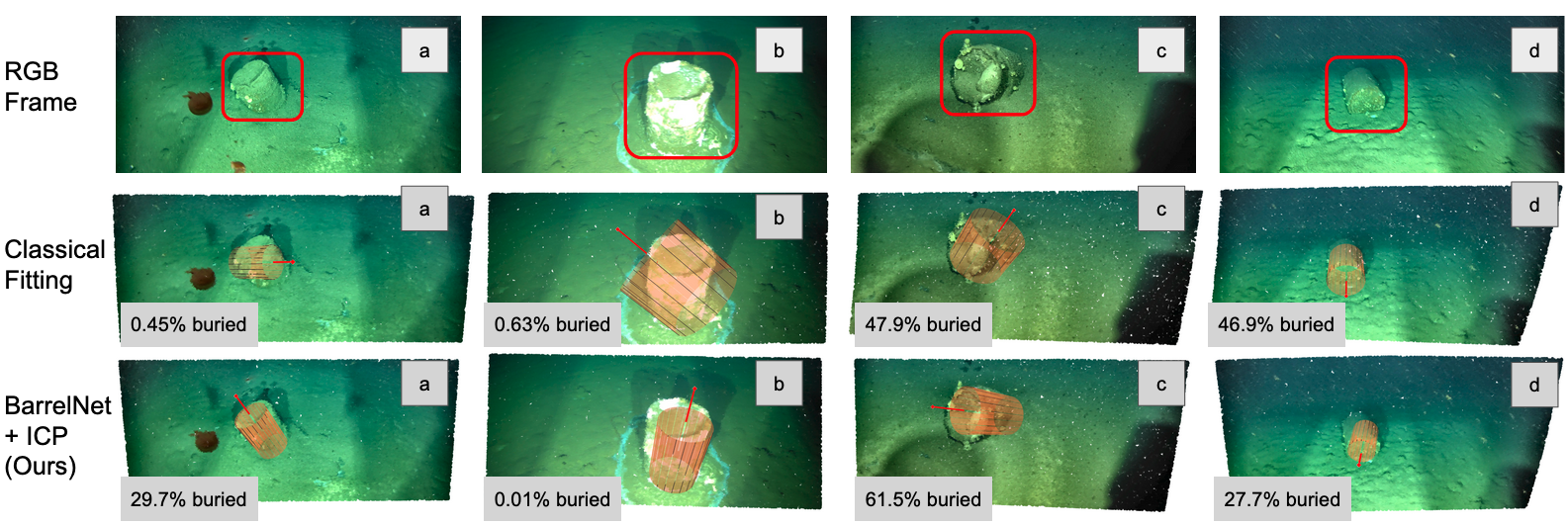}
    \caption{Results on 4 different seabed barrel scenes (top row) obtained at a water depth of 850m in the San Pedro Basin, CA. As a qualitative metric, we overlay the fitted cylinders (orange) on the reconstructed DUSt3R point clouds. The barrel axis vectors are shown as red rays. Classical least squares method (middle row) produces inconsistent results due to MLESAC randomness, except for (d). 
    For our method (bottom row), despite slight mismatches (c,d) the barrel axis vectors are largely oriented correctly w.r.t. the ocean floor.}
    \label{fig:cyl-fit-results}
\end{figure*}

We evaluate our model quantitatively on a test set of synthetically generated point clouds. We evaluate burial fraction error and cosine similarity $(|\langle \vec{\mathbf{n}}_{\text{gt}}, \vec{\mathbf{n}}_{\text{pred}} \rangle |)$ between the predicted and true barrel axis as numerical metrics for our  benchmarks. We also show qualitative visual evaluations of the predictions from our approach on the deep-sea ROV footage from a survey in the San Pedro Basin, CA \cite{merrifield_wide-area_2023}.

\subsection{Synthetic Data Evaluation}

We evaluate the BarrelNet model on a set of 1,000 randomly oriented cylinders, each of which are occluded by 20 random camera poses (20,000 different point clouds). In addition to evaluating BarrelNet, we also test a classical cylinder fitting method which estimates the axis based on least square fitting to the surface normals \cite{vosselman_recognising_2003} and filters outliers based on MLESAC \cite{torr_mlesac_2000}.

Due to the incomplete nature of the video footage of barrels, most point clouds reconstructed from this footage do not capture a full 360$^\circ$ view of each barrel. This is because the ROV typically does not perform a complete circular survey around each barrel during seabed missions. Thus, we both train and test BarrelNet on noisy cylinder point clouds, where we exclude points based on visual occlusion, so the only points available are above the seafloor ($z\geq0$) and visible to a set camera view.

\textbf{Results:} We observed that occluded and noisy cylinders with caps have a strongly negative impact on the accuracy of classical fitting methods. BarrelNet on the other hand, is remarkably robust to occlusion effects in comparison to classical least squares fitting based methods. Quantitatively, we observed a mean cosine similarity (higher is better) of 0.617, and burial fraction error (lower is better) of 25.8\% for the classical method. For BarrelNet we observed mean cosine similarity of \textbf{0.997} and mean burial fraction error of \textbf{1.9\%}.

\subsection{ROV Footage Results}

Due to the challenging environment of the dumpsite, obtaining precise ground truth statistics for barrel orientation and burial fraction is extremely difficult. Thus, we rely on qualitative assessments of workflow performance such as visualizing the cylinder meshes instantiated in the predicted pose onto the input images. We show some examples in \autoref{fig:cyl-fit-results}.

The four different scenes in \autoref{fig:cyl-fit-results} were found to be amenable for good quality 3D reconstruction from DUSt3R. Classical least squares is observed to produce inconsistent results, as MLESAC fits a different cylinder for every prediction. Aside from barrel (d), $\vec{\mathbf{n}}$ predictions are influenced by camera occlusions and the presence of cylinder cap points.

On the contrary, BarrelNet and ICP estimate visually consistent barrel poses and burial percentages barring slight mismatches in the barrel centroid locations. Prior knowledge of barrel details, such as number of visible reinforcement rings helps in assessing the burial estimation quality. For example, nearly vertical barrels, such as (a) and (b) in \autoref{fig:cyl-fit-results} are almost certainly buried $<50\%$. Based on these qualitative priors we conclude that classical fitting burial percentages in (a,d) have more error.

Despite the fitted cylinders having some minor errors, this shows there are several avenues of improvement to our workflow. Since a 3D reconstruction model like DUSt3R lacks training data with underwater footage, synthetic scenes simulating underwater imaging conditions can be used as ground truth reconstructions are difficult to obtain. Alternatively, barrel poses can be estimated straight from images using a model like FoundationPose \cite{wen_foundationpose_2024}, although fundamental changes would have to be made to account for large variations in the degradation state of each barrel. 

\section{Conclusion}
We introduce a novel workflow for estimating the pose and burial fraction of barrels on the ocean floor. Through extensive synthetic evaluations, our computer vision-based estimates of the burial fraction and barrel pose show significant promise. Accurate estimation of these quantities is crucial for assessing the impact of these barrels on the surrounding ocean environment and sedimentation rates. This evaluation will help determine necessary steps for monitoring the ecological impact of containerized waste in marine environments. We plan to extend our method to generalize across a broader range of object types in the future.

\bibliography{references}
\bibliographystyle{IEEEtran}

\end{document}